\documentclass[10pt,twocolumn,letterpaper]{article}
\usepackage[utf8]{inputenc}
\usepackage{iccv}
\usepackage{times}
\usepackage[inline]{enumitem}
\usepackage[export]{adjustbox}
\usepackage{amsmath}
\usepackage{amssymb}
\usepackage{bm}  %
\usepackage{multirow}
\usepackage{booktabs}
\usepackage{calc}
\usepackage{chngcntr}  %
\usepackage{silence}
\usepackage{fixltx2e}
\usepackage{dblfloatfix}  %
\usepackage{framed}
\usepackage{graphicx}
\usepackage{microtype}
\usepackage{pifont}%
\usepackage{placeins}  %
\usepackage{xfrac}  %
\usepackage{subcaption}
\usepackage{makecell}
\usepackage{tikz}
\usetikzlibrary{arrows.meta,positioning,fit,backgrounds,calc}
\usepackage[pagebackref=true,breaklinks=true,colorlinks,bookmarks=false]{hyperref} %
\usepackage{cleveref} %

\usepackage{multibib}
\newcites{suppl}{Supplemental references}%

\makeatletter
\renewcommand\@makefntext[1]{%
  \noindent\makebox[1em][r]{\@makefnmark}#1}
\makeatother

\let\vec\bm

\newcommand{\inaug}{IN-aug\xspace}

\newcommand{\bx}{\vec{x}}

\newcommand{\be}{\vec{e}}
\newcommand{\bw}{\vec{w}}

\newcommand{\bW}{\vec{W}}

\newcommand{\lclass}{\ell^\textrm{class}}
\newcommand{\lret}{\ell^\textrm{retr}}

\makeatletter
\renewcommand{\paragraph}{%
  \@startsection{paragraph}{4}%
  {\z@}{0.4em}{-1em}%
  {\normalfont\normalsize\bfseries}%
}

\setlist[itemize]{%
labelsep=5pt,%
labelindent=0.4\parindent,%
itemindent=0pt,%
leftmargin=*,%
itemsep=-4pt, 
}

\makeatother

\definecolor{darkgreen}{RGB}{0, 140, 0}
\definecolor{antiquefuchsia}{rgb}{0.57, 0.36, 0.51}
\definecolor{auburn}{rgb}{0.43, 0.21, 0.1}

\newif\ifarxiv
\global\arxivtrue

\iccvfinalcopy
\ificcvfinal\fi
\def\iccvPaperID{5147}

\ifarxiv\fi

\title{MultiGrain: a unified image embedding for classes and instances}
\author{\begin{tabular}{ccccc}
  Maxim Berman\thanks{Did this work during an internship at Facebook AI Research.} & Herv\'e J\'egou & Andrea Vedaldi & Iasonas Kokkinos & Matthijs Douze \\
  ESAT--PSI, KU Leuven & \multicolumn{4}{c}{Facebook AI Research} \\
\end{tabular}}

\ificcvfinal
\hypersetup{  %
pdftitle={MultiGrain: a unified image embedding for classes and instances},
pdfsubject={},
pdfauthor={Maxim Berman, Herve Jegou, Andrea Vedaldi, Iasonas Kokkinos, Matthijs Douze} %
pdfkeywords={}
}
\fi

\makeatletter
\let\inserttitle\@title
\makeatother

\begin{document}
\maketitle
\begin{abstract}
MultiGrain is a network architecture producing compact vector representations that are suited both for image classification and particular object retrieval. 
It builds on a standard classification trunk.
The top of the network produces an embedding containing coarse and fine-grained information, so that images can be recognized based on the object class, particular object, or if they are distorted copies.
Our joint training is simple: we minimize a  cross-entropy loss for classification and a ranking loss 
that determines if two images are identical up to data augmentation, with no need for additional labels. 
A key component of MultiGrain is a pooling layer that takes advantage of high-resolution images with a network trained at a lower resolution. 

When fed to a linear classifier, the learned embeddings provide state-of-the-art classification accuracy. 
For instance, we obtain 79.4\% top-1 accuracy with a ResNet-50 learned on Imagenet, which is a +1.8\% absolute improvement over the AutoAugment method. 
When compared with the cosine similarity, the same embeddings perform on par with the state-of-the-art for image retrieval at moderate resolutions. 

\end{abstract}

\section{Introduction}\label{sec:intro}

Image recognition is central to computer vision, with dozens of new approaches being proposed every year, each optimized for particular aspects of the problem.
From coarse to fine, we may distinguish the recognition of
(a) classes, where one looks for a certain type of object regardless of intra-class variations,
(b) instances, where one looks for a particular object despite changes in the viewing conditions,  and
(c) copies, where one looks for a copy of a specific image despite edits.
While these problems are in many ways similar, the standard practice is to use specialized, and thus incompatible, image representations for each case.

Specialized representations may be accurate, but constitute a significant bottleneck in some applications.
Consider for example image retrieval, where the goal is to match a query image to a large database of other images.
Very often one would like to search the same database with multiple granularities, by matching the query by class, instance, or copy.
The performance of an image retrieval system depends primarily on the \emph{image embeddings} it uses. 
These strike a trade-off between database size, matching and indexing speed, and retrieval accuracy.
Adopting multiple embeddings, narrowly optimized for each type of query, means multiplying the resource usage.

\begin{figure}
\includegraphics[width=1.\linewidth]{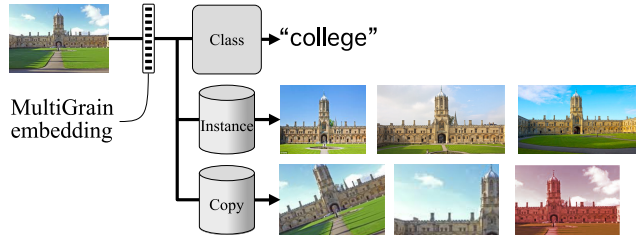}
\vspace{-10pt}
\caption{Our goal is to extract an image descriptor incorporating different levels of granularity, so that we can solve, in particular, classification and particular object recognition tasks: The descriptor is either fed to a linear classifier, or directly compared with cosine similarity. 
\label{fig:splash}} 
\vspace{-10pt}
\end{figure}

In this paper we present a new  representation, MultiGrain, that can achieve the three tasks together, regardless of differences in their semantic granularity, see \cref{fig:splash}.
We learn MultiGrain by jointly training an image embedding for multiple tasks.
The resulting representation is compact and outperforms narrowly-trained embeddings.

Instance retrieval has a wide range of industrial applications, including detection of copyrighted images and exemplar-based recognition of unseen objects.
In settings where billion of images have to be treated, %
it is of interest to obtain image embeddings suitable for more than one recognition task.
For instance, an image storage platform is likely to perform some classification of the input images, aside from detecting copies or instances of the same object.
An embedding relevant to all these tasks advantageously reduces both the computing time per image and storage space.

In this perspective, convolutional neural networks (CNNs) trained only for classification already go a long way towards universal features extractors. %
The fact that we can learn image embeddings that are simultaneously good for classification and instance retrieval is surprising but not contradictory.
In fact, there is a logical dependence between the tasks: images that contain the same instance also contain, by definition, the same class; and copied images contain the same instance.
This is in contrast to multi-task settings where tasks are in competition and are thus difficult to combine.
Instead, both class, instance, and copy congruency lead to embeddings that should be close in feature space.
Still, the degree of similarity is different in the different cases, with classification requiring more invariance to appearance variations and copy detection sensitivity to small image details.

In order to learn an image representation that satisfies the different trade-offs, we start from an existing image classification network.
We use a generalized mean layer %
that converts a spatial activation map to a fixed-size vector.
Most importantly, we show that it is an effective way to learn an architecture that can adapt to different resolutions \emph{at test time}, and offer higher accuracies. This circumvents the massive engineering and computational effort needed to learn networks for larger input resolutions~\cite{He2016IdentityMI}

The joint training of classification and instance recognition objectives is based on cross-entropy and contrastive losses, respectively.
Remarkably, instance recognition is learned for free, \emph{without using labels specific to instance recognition or image retrieval}: we simply use the identity of the images as labels, and data augmentation as a way to generate different versions of each image. %

In summary, our main contributions are as follows: 
\begin{itemize}
    \item We introduce the MultiGrain architecture, which outputs an image embedding incorporating different levels of granularity. Our dual classification+instance objective improves the classification accuracy on its own. %
    \item We show that part of this gain is due to the batching strategy, where each batch contains repeated instances of its images with different data augmentations for the purpose of the retrieval loss; %
    \item We incorporate a pooling layer inspired by image retrieval. It provides a significant boost in classification accuracy when provided with high-resolution images.
\end{itemize}
Overall, our architecture offers competing performance both for classification and image retrieval. Noticeably, we report a significant boost in accuracy on Imagenet with a ResNet-50 network over the state of the art.

The paper is organized as follows. \Cref{sec:related} introduces related works. \Cref{sec:arch} introduces our architecture, the training procedure and explains how we adapt the resolution at test time. \Cref{sec:experiments} reports the main experiments. %

\section{Related work}
\label{sec:related}
\paragraph{Image classification.} 
Most computer vision architectures designed for a wide range of tasks leverage a trunk architecture initially designed for classification, such as Residual networks~\cite{he16resnet}. An improvement on the trunk architecture eventually translates to better accuracies in other tasks~\cite{he2017mask}, as shown on the detection task of the LSVRC'15 challenge. While recent architectures~\cite{hu2018squeeze,huang2017densely,Xie2017AggregatedRT} have exhibited some additional gains, other lines of research have been investigated successfully. For instance, a recent trend~\cite{mahajan2018exploring} is to train high capacity networks by leveraging much larger training sets of of weakly annotated data. 
To our knowledge, the state of the art on Imagenet ILSVRC 2012 benchmark for a model learned from scratch on Imagenet train data only is currently hold by the gigantic AmoebaNet-B architecture~\cite{huang2018gpipe} (557M parameters), which takes 480x480 images as input. 

In our paper, we choose ResNet-50~\cite{he16resnet} (25.6M parameters), as this architecture is adopted in the literature in many works both on image classification and instance retrieval. 

\paragraph{Image search: from local features to CNN.} 
``Image search'' is a generic retrieval task that is usually associated with and evaluated for more specific problems such as landmark recognition~\cite{Jegou2008HammingEA,Philbin07}, particular object recognition~\cite{nister2006scalable} or copy detection~\cite{Douze2009EvaluationOG}, for which the objective is to find the images most similar to the query in a large image collection. 
In this paper ``image retrieval'' will refer to instance-level retrieval, where object instances are as broad as possible, \ie not restricted to buildings, as in the Oxford/Paris benchmark.
Effective systems for image retrieval rely on accurate image descriptors.  
Typically, a query image is described by an embedding vector, and the task amounts to searching the nearest neighbors of this vector in the embedding space. Possible improvement include refinement steps such as geometric verification \cite{Philbin07}, query expansion~\cite{chum2007total,TOLIAS20143466}, or database-side pre-processing or augmentation \cite{tolias2016image,turcot2009better}.

Local image descriptors are traditionally aggregated to global image descriptors suited for matching in an inverted database, as in the seminal bag-of-words model~\cite{Sivic2003VideoGA}.
After the emergence of convolutional neural networks (CNNs) for large-scale classification on ImageNet~\cite{krizhevsky2012imagenet, ILSVRC15}, it has become apparent that CNNs trained on classification datasets are very competitive image feature extractors for various vision tasks, including instance retrieval~\cite{babenko2014neural,gong2014multi,Razavian2014CNNFO}.

\paragraph{Specific architectures for particular object retrieval}\hspace{-1em} are built upon a regular classification trunk, and modified so the pooling stage gives more spatial locality in order to cope with small objects and clutter. 
For instance, a competitive baseline for instance retrieval on various datasets is the R-MAC image descriptor~\cite{Tolias2015ParticularOR}. It aggregates regionally pooled features extracted from a CNN. %
The authors show that this specialized pooling combined with PCA whitening~\cite{jegou2012negative} leads to efficient many-to-many comparisons between image regions, highly beneficial to image retrieval.
Gordo~et~al.~\cite{Gordo2016DeepIR, Gordo2017EndtoEndLO} show that fine-tuning these regionally-aggregated representations end-to-end on an external image retrieval dataset using a ranking loss yields significant improvements for instance retrieval. 

Radenovi{\'c}~\etal~\cite{radenovic2018fine} show that R-MAC pooling is advantageously replaced by a generalized mean pooling (see \cref{sec:p-pooling}), which is a spatial pooling of the features exponentiated to an exponent $p$ over the whole image. The exponentiation localizes the features on the point of interests in the image, replacing regional aggregation in R-MAC. %

\paragraph{Multi-task training}\hspace{-1em}
is an active area of research~\cite{Kokkinos2017UberNetTA,Zamir2018TaskonomyDT}, motivated by the observation that deep neural networks are transferable to a wide range of vision tasks~\cite{Razavian2014CNNFO}. 
Moreover trained deep neural networks exhibit a high level of compressibility~\cite{han2015deep}. 
In some cases, sharing the capacity of neural networks between different tasks through shared parameters  
helps the learning by allowing complementary training among datasets and low-level features. 
Despite some successes with multi-task networks for vision such as UberNet~\cite{Kokkinos2017UberNetTA}, the design and training of multi-task networks still involve numerous heuristics. 
Ongoing lines of work include finding the right architecture for an efficient sharing of parameters~\cite{rebuffi2018efficient}, and finding the right optimization parameters for such networks in order to depart from the traditional setting of single-task single-dataset end-to-end gradient descent, and efficiently weight the gradients in order to obtain a well-performing network in all tasks~\cite{guo2018dynamic}. 

\paragraph{Data augmentation}\hspace{-1em} is a cornerstone of the training in large-scale vision applications~\cite{krizhevsky2012imagenet}, which improves generalization and reduces over-fitting. 
In a stochastic gradient descent (SGD) optimization setting, we show that including multiple data-augmented instances of the same image in one optimization batch, rather than having only distinct images in the batch, significantly enhances the effect of data-augmentations and improve the generalization of the network. 
A related batch augmented (BA) sampling strategy was concurrently introduced by Hoffer et al.~\cite{2019arXiv190109335H}. 
When augmenting the size of the batches in a large-scale distributed optimization of a neural network, they show that filling these bigger batches with data-augmented copies of the image in the batch yields better generalization performance, and uses computing resources more efficiently through reduced data processing time. 
As discussed in~\cref{sec:data-augmented-batches} and highlighted in our classification results (\cref{sec:classif-results}), we show that a gain in performance under this sampling scheme is obtained~\emph{using the same batch size}, \ie, with a lower number of distinct images per batch. 
We consider this scheme of repeated augmentations (RA) within the batch as a way to boost the effect of data augmentation over the course of the optimization. 
Our results indicate that RA is a technique of general interest, beyond large-scale distributed training applications, for improving the generalization of neural networks.

\section{Architecture design}\label{sec:arch}
Our goal is to develop a convolutional neural network that is suitable for both image classification and instance retrieval.
In the current best practices, the architectures and training procedures used for class and instance recognition differ in a significant manner.

\begin{figure}
\centering
\includegraphics[width=0.9\linewidth]{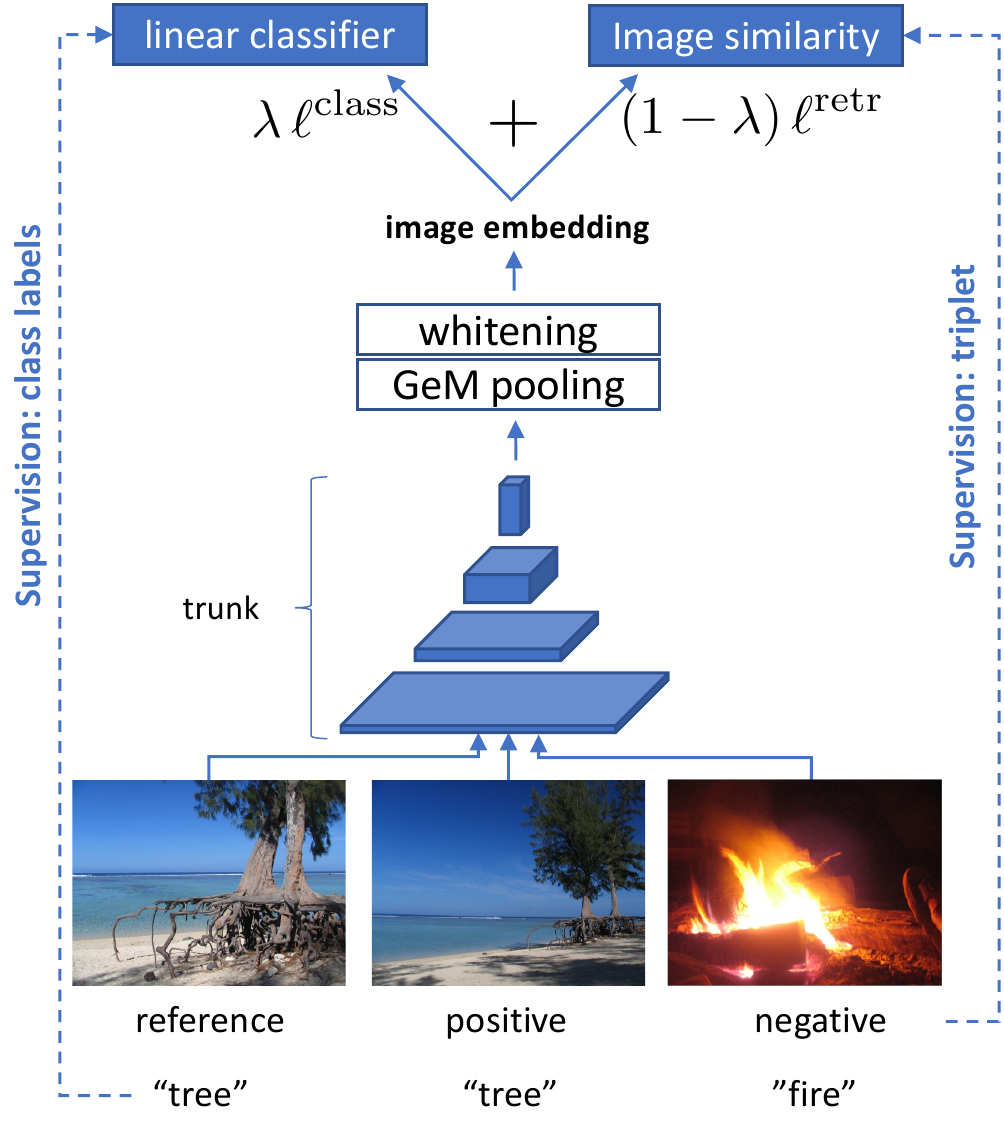}
\caption{\label{fig:3arch}
 Overview of our Multigrain architecture.} %
 \vspace{-10pt}
\end{figure}

This section describes such technical differences, summarized in \cref{tab:diff_classif_instance}, together with our solutions to bridge them. 
This leads us to a unified architecture, shown in \cref{fig:3arch}, that we jointly train for both tasks in an end-to-end manner.

\begin{table}
\centering
\caption{\label{tab:diff_classif_instance}
Differences between classification and image retrieval:
Retrieval architectures incorporate a final pooling layer that is regionalized (RMAC) or magnifies activations (GeM). The triplet loss requires a batching strategy with pairs of matching images. %
}
\vspace{-7pt}
{\small
\begin{tabular}{@{}lc@{\hspace{5pt}}c@{}}
\toprule
                  & classification       & retrieval \\ 
\midrule 
spatial pooling   & avg. pooling         & RMAC~\cite{TOLIAS20143466} or GeM~\cite{radenovic2018fine} \\
loss              & cross-entropy        & triplet~\cite{Gordo2016DeepIR}  \\
batch sampling    & diverse              & similar images in batch  \\
whitening         & no                   & yes  \\
resolution        & low ($224^2$--$300^2$)           & high ($800$--$1$k$\times$scaled) \\
\bottomrule
\end{tabular}}

\vspace{-7pt}
\end{table}

\subsection{Spatial pooling operators\label{sec:p-pooling}}
This section considers the final, global spatial pooling layer.
Local pooling operators, usually max pooling, are found throughout the layers of most convolutional networks to achieve local invariance to small translations. 
By contrast, global spatial pooling converts a 3D tensor of activations produced by a convolutional trunk to a vector. 

\paragraph{Classification.}  
In early models such as LeNet-5~\cite{lecun1989backpropagation} or AlexNet~\cite{krizhevsky2012imagenet}, the final spatial pooling is just a linearization of the activation map.
It is therefore sensitive to the absolute location. 
Recent architectures such as ResNet and DenseNet employ average pooling, which is permutation invariant and hence offers a more global translation invariance. 

\paragraph{Image retrieval}\hspace{-1em} requires more localized geometric information: particular objects or landmarks are visually more similar, but the task suffers more from clutter, and a given query image has no specific training data devoted to it. This is why the pooling operator tries to favor more locality. Next we discuss the generalized mean pooling operator. 

Let $\bx\in\mathbb{R}^{C\times W \times H}$ be the feature tensor computed by a convolutional neural network for a given image, where $C$ is the number of feature channels and $H$ and $W$ are the height and width of the map, respectively. 
We denote by $u\in\Omega=\{1,\dots,H\}\times\{1,\dots,W\}$ a ``pixel'' in the map, by $c$ the channel, and by $x_{cu}$ the corresponding tensor element: $\bx=[x_{cu}]_{c=1..C,u\in\Omega}$.
The generalized mean pooling (GeM) layer computes the generalized mean of each channel in a tensor.
Formally, the GeM embedding is given by
\begin{equation}\label{eq:gem}
 \be
 =
 \left[
 \Big(\frac{1}{|\Omega|}
 \sum_{u\in \Omega}x^p_{cu}
\Big)^\frac{1}{p}
 \right]_{c=1..C}
\end{equation}
where $p > 0$ is a parameter. Setting this exponent as $p>1$ increases the contrast of the pooled feature map and focuses on the salient features of the image~\cite{Bo2009EfficientMK,Boureau2010ATA,dollar2009integral}. 
GeM is a generalization of the average pooling commonly used in classification networks ($p=1$) and of spatial max-pooling layer ($p=\infty$). 
It is employed in the original R-MAC as an approximation of max pooling~\cite{dollar2009integral}, yet only recently~\cite{radenovic2018fine} it was shown to be competitive on its own with R-MAC for image retrieval.

To the best of our knowledge, this paper is the first to apply and evaluate GeM pooling in an image classification setting.
More importantly, we show later in this paper that \emph{adjusting the exponent is an effective way to change the input image resolution between train and test time} for all tasks, which explains why image retrieval has benefited from it considering that this task employs higher-resolution images.

\subsection{Training objective\label{sec:losses}}

In order to combine the classification and retrieval tasks, we use a joint objective function composed of a classification loss and an instance retrieval loss. 
The two-branch architecture is illustrated in ~\cref{fig:3arch} and detailed next.

\paragraph{Classification loss.}

For classification, we adopt the standard cross-entropy loss.
Formally, let $\be_i\in\mathbb{R}^d$ be the embedding computed by the deep network for image $i$, $\bw_c \in \mathbb{R}^{d}$ the parameters of a linear classifier for class $c \in \{1,\dots,C\}$, and $y_i$ be the ground-truth class for that image.
Then
\begin{equation}\label{eq:crossent}
\lclass(\be_i,\bW,y_i)
=
- \langle \bw_{y_i}, \be_i \rangle
+ \log \sum_{c=1}^C
\exp \langle\bw_c, \be_i \rangle,
\end{equation}
where $\bW=[\bw_c]_{c=1..C}$.
We omit it for simplicity, but by adding a constant channel to the feature vector, the  bias of the classification layer is incorporated in its weight matrix.

\paragraph{Retrieval loss.}

For image retrieval, the embeddings of two matching images (a positive pair) should have distances smaller than embeddings of non-matching images (a negative pair). 
This can be enforced in two ways. 
The contrastive loss~\cite{hadsell2006dimensionality} requires distances between positive pairs to be smaller than a threshold, and distances between negative pairs to be greater.
The triplet loss instead requires an image to be closer to a positive sibling than to a negative sibling~\cite{schroff2015facenet}, which is relative property of image triplets.
These losses requires adjusting multiple parameters, including how pairs and triplets are sampled. These parameters are sometimes hard to tune, especially for the triplet loss.

Wu~et~al.~\cite{wu2017sampling} proposed an effective method that addresses these difficulties.
Given a batch of images, they re-normalize their embeddings to the unit sphere, sample negative pairs as a function of the embedding similarity, and use those pairs in a margin loss, a variant of contrastive loss that shares some of the benefits of the triplet loss.

In more detail, given images $i,j\in\mathcal{B}$ in a batch with embeddings $\be_i,\be_j\in\mathbb{R}^d$, the margin loss is expressed as
\begin{equation}\label{eq:margin}
  \lret(\be_i, \be_j, \beta, y_{ij})
  = \max\{0, \alpha + y_{ij}(D(\be_i, \be_j) - \beta)\}
\end{equation}
where
$
    D(\be_i, \be_j) = \left\| \sfrac{\be_i}{\|\be_i\|} - \sfrac{\be_j}{\|\be_j\|}\right\|
$
is the Euclidean distance between the normalized embeddings, the label $y_{ij}$ is equal to 1 if the two images match and $-1$ otherwise, $\alpha > 0$ the margin (a constant hyper-parameter), and $\beta > 0$ is a parameter (learned during training together with the model parameters), controlling the volume of the embedding space occupied embedding vectors.
Due to the normalization, $D(\be_i, \be_j)$ is equivalent to a cosine similarity, which, up to whitening (\cref{sec:whiten}), is also used in retrieval.

Loss~\eqref{eq:margin} is computed on a subset of positive and negative pairs $(i,j)\in \mathcal{B}^2$ selected with the sampling~\cite{wu2017sampling} 
\begin{equation}
\begin{aligned}
\mathcal{P}_+(\mathcal{B})
&= \{ (i,j) \in \mathcal{B}^2: y_{ij} = 1\},
\\
\mathcal{P}_-(\mathcal{B})
&=
  \{ (i,j^*) : (i,j) \in \mathcal{P}_+(\mathcal{B}), 
  j^* \sim p(\cdot|i)
  \},
\\
\mathcal{P}(\mathcal{B})&=\mathcal{P}_+(\mathcal{B})\cup\mathcal{P}_-(\mathcal{B}),
\end{aligned}
\end{equation}
where the conditional probability of choosing a negative $j$ for image $i$ is 
$
 p(j|i) \propto \min\{\tau, q^{-1}(D(\be_i,\be_j))\} 
 \cdot \mathbf{1}_{\{y_{ij}=-1\}},
$
where $\tau > 0$ is a parameter and $q(z) \propto z^{d-2}(1 - z^2/4)^{\frac{d-3}{2}}$ is a PDF that depends on the embedding dimension $d$.

The use of distance weighted-sampling with margin loss is very suited to our joint training setting: this framework tolerates relatively small batch sizes ($|\mathcal{B}|\sim 80$ to $120$ instances) while requiring only a small amount of positives images (3 to 5) of each instance in the batch, without the need for elaborate parameter tuning or offline sampling.

\paragraph{Joint loss and architecture.}

The joint loss is a combination
of classification and retrieval loss weighted by a factor $\lambda \in [0, 1]$.
For a batch $\mathcal{B}$ of images, the joint loss writes as
\begin{equation}\label{eq:joint}
\frac{\lambda}{|\mathcal{B}|}\cdot\sum_{i\in\mathcal{B}}
\lclass(\be_i,\bw,y_i)
+
\frac{1-\lambda}{|\mathcal{P}(\mathcal{B})|}
\cdot\hspace{-1em}
\sum_{\raisebox{-3pt}{
$\scriptstyle(i,j)\in\mathcal{P}(\mathcal{B})$
}}
\hspace{-1em}
\lret(\be_i, \be_j, \beta, y_{ij}),
\end{equation}
\ie, losses are normalized by the number of items in the corresponding summations.

\subsection{Batching with repeated augmentation (RA)\label{sec:data-augmented-batches}}

Here, we propose to use only a training dataset for image classification, and train instance recognition via data augmentation.
The rationale is that data augmentation produces another image that contains the same object instance.
This approach does not require more annotation beyond the standard classification set.

We introduce a new sampling scheme for training with SGD and data augmentation, which we refer to as~\emph{repeated augmentations}.
In RA we form an image batch $\mathcal{B}$ by sampling $\left\lceil|\mathcal{B}|/m\right\rceil$ different images from the dataset, and transform them up to $m$ times by a set of data augmentations to fill the batch. 
Thus, the instance level ground-truth $y_{ij}=+1$ iff images $i$ and $j$ are two augmented versions of the same training image.
The key difference with the standard sampling scheme in SGD is that samples are not independent, as augmented versions of the same image are highly correlated. 
While this strategy reduces the performance if the batch size is small, for larger batch sizes RA outperforms the standard i.i.d.~scheme -- while using the same batch size and learning rate for both schemes.
This is different from the observation of~\cite{2019arXiv190109335H}, who also consider repeated samples in a batch, but simultaneously increase the size of the latter.

We conjecture that the benefit of correlated RA samples is to facilitate learning features that are invariant to the only difference between the repeated images --- the augmentations.
By comparison, with standard SGD sampling, two versions of the same image are seen only in different epochs. 
A study of an idealized problem illustrates this phenomenon in the supplementary material~\ref{sec:suppl-data-augmented-toy}.

\subsection{PCA whitening}\label{sec:whiten}

In order to transfer features learned via data augmentation to standard retrieval datasets, we apply a step of PCA whitening, in accordance
with previous works in image retrieval~\cite{Gordo2017EndtoEndLO,jegou2012negative}.
The Euclidean distance between transformed features is equivalent to the Mahalanobis distance between the input descriptors.
This is done after training the network, using an external dataset of unlabelled images.

The effect of PCA whitening can be undone in the parameters of the classification layer, so that the whitened embeddings can be used for both classification and instance retrieval. In detail, let $\be$ be an image embedding vector and $\bw_c$ the weight vector for class $c$, such that $\langle\bw_c,\be\rangle$ are the outputs of the classifier as in~\cref{eq:crossent}.
The whitening operation $\Phi$ can be written as~\cite{Gordo2017EndtoEndLO}
$
    \Phi(\be) = S \left(\frac{\be}{\|\be\|} - \bm{\mu}\right)
$
given the whitening matrix $S$ and centering vector $\bm{\mu}$; hence
\begin{equation*}\label{eq:adapted_classifier}
    \langle\bw_c,\be\rangle = 
    \langle\bw_c,\Phi^{-1}(\Phi(\be))\rangle =
    \|\be\| \left( \langle\bw_c', \Phi(\be)\rangle + b_c' \right) 
\end{equation*}
where 
$
\bw_c' =  
S^{-1 \top} \bw_c
$
and
$
b_c' = \langle\bw_c, \mu \rangle
$
are the modified weight and bias for class $c$.
We observed that inducing decorrelation via a loss~\cite{Cogswell2016ReducingOI} is insufficient to ensure that features generalize well, which concurs with prior works~\cite{Gordo2017EndtoEndLO,radenovic2018fine}.

\subsection{Input sizes\label{sec:input-size}}

The standard practice in image classification is to resize and center-crop input images to a relatively low resolution, e.g.~$224\times 224$ pixels~\cite{krizhevsky2012imagenet}. 
The benefits are a smaller memory footprint, faster inference, and the possibility of batching the inputs if they are cropped to a common size. On the other hand, image retrieval is typically dependent on finer details in the images, as an instance can be seen under a variety of scales, and cover only a small amount of pixels.
The currently best-performing feature extractors for image retrieval therefore commonly use input sizes of $800$~\cite{Gordo2017EndtoEndLO} or $1024$~\cite{radenovic2018fine} pixels for the largest side, without cropping the image to a square.
This is impractical for end-to-end training of a joint classification and retrieval network.

Instead, we train our architecture at the standard $224\times 224$ resolution, and use larger input resolutions at test time only. 
This is possible due to a key advantage of our architecture:
a network trained with a pooling exponent $p$ and resolution $s$ can be evaluated at a larger resolution $s^*>s$ using a larger pooling exponent $p^*>p$, see our validation in \cref{sec:classif-results}. 

\label{sec:expanding-resolution}
\paragraph{Proxy task for cross-validation of $p^*$.} In order to select the exponent $p^*$, suitable for all tasks, we create a synthetic retrieval task {\bf \inaug} in between classification and retrieval. We sample $2,\!000$ images from the training set of ImageNet, $2$ per class, and create 5 augmented copies of each of them, using the ``full'' data augmentation described before. 

We evaluate the retrieval accuracy on \inaug in a fashion similar to UKBench~\cite{nister2006scalable}, with an accuracy ranging from 0 to 5 depending measuring how many of the first 5 augmentations are ranked in top 5 positions. 
We pick the best-performing $p^* \in \{1, 2,\ldots, 10\}$ on \inaug, 
which provides the following choices as a function of $\lambda$ and $s^*$:
\begin{equation*}
{\small
\begin{tabular}{c|lrrr}
\toprule
$\lambda$ & $s^*=$ & $224$ & $500$ & $800$ \\
\midrule
$1$   & $p^*=$ & $3$ & $4$ & $4$\\
$0.5$ & $p^*=$ & $3$ & $4$ & $5$\\
\bottomrule
\end{tabular}}
\end{equation*}

The optimal $p^*$ obtained on \inaug provides a trade-off between retrieval and classification. Experimentally, we observed that other choices are suitable for setting this parameter: %
fine-tuning the parameter $p^*$ alone using training inputs at a given resolution by back-propagation of the cross-entropy loss provides similar results and values of $p^*$.

\section{Experiments and Results\label{sec:experiments}}

After presenting the datasets, we provide a parametric study and our results in image classification and retrieval.

\subsection{Experimental settings\label{sec:experimental-settings}}
\paragraph{Base architecture and training settings.}

The convolutional trunk is ResNet-50~\cite{he16resnet}. 
SGD starts with a learning rate of $0.2$ which is reduced tenfold at epochs $30, 60, 90$ for a total of $120$ epochs (a standard setting~\cite{paszke2017automatic}). 
The batch size $|\mathcal{B}|$ is set to $512$ and an epoch is defined as a fixed number of $T=5005$ iterations. With uniform batch sampling, one epoch corresponds to two passes over the training set; with RA and $m=3$, one epoch corresponds to $\sim 2/3$ of the images of the training set. 
All classification baselines are trained using this longer schedule for a fair comparison. 

\paragraph{Data augmentation.} We use standard flips, random resized crops~\cite{howard2013some}, random lighting noise and a color jittering of brightness, contrast and saturation~\cite{krizhevsky2012imagenet,howard2013some}. 
We refer to this set of augmentations as ``full'', see details in supplemental~\ref{sec:full-data-augment}. 
As indicated in~\cref{tab:classifres} our network reaches $76.2$\% top-1 validation error under our chosen schedule and data augmentation when trained with cross-entropy alone and uniform batch sampling. 
This figure is on the high end of accuracies reported for the ResNet-50 network~\cite{goyal2017accurate,he16resnet} without specially-crafted regularization terms~\cite{zhang2018mixup} or data augmentations~\cite{cubuk2018autoaugment}.

\begin{figure}
{\small
\begin{minipage}{0.47\columnwidth}
\centering 
Input image \medskip

\includegraphics[width=\columnwidth]{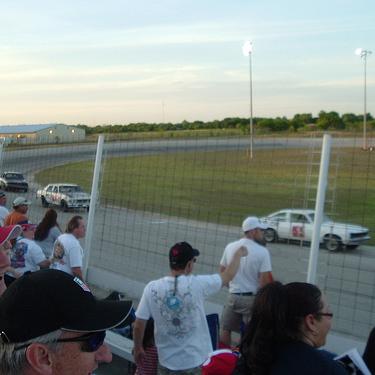} 

\end{minipage}%
\hfill
\begin{minipage}{0.5\columnwidth}
\centering
resolution $s^*=224$ \\

$p^* = 1$ \hspace{1cm} $p^* = 3$ \\

\includegraphics[width=0.8\columnwidth]{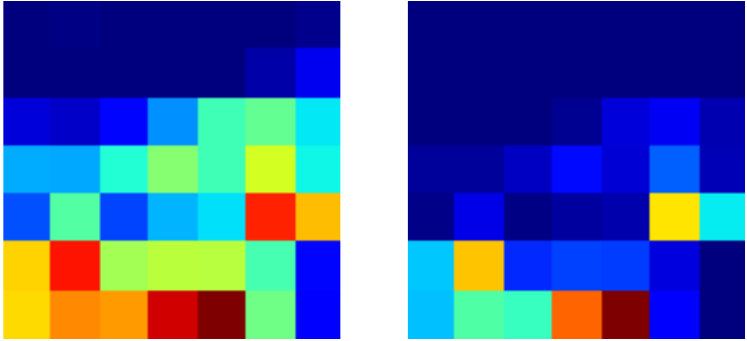}

full resolution \\

$p^* = 1$ \hspace{1cm} $p^* = 3$ \\

\includegraphics[width=0.8\columnwidth]{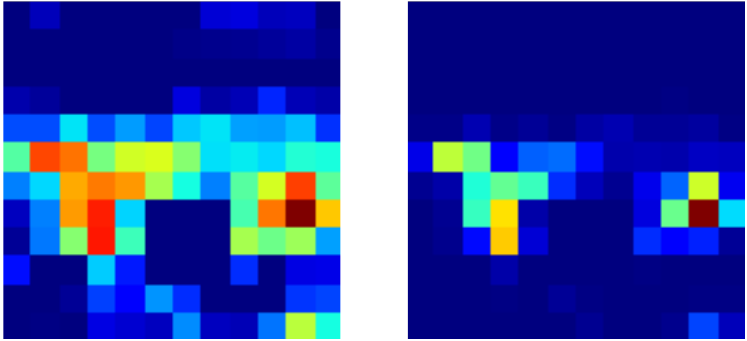}
\end{minipage}}
\vspace{-5pt}

\caption{\label{fig:heatmap}
    An off-the-shelf ResNet-50 reacts strongly on channel 909 of the last activation map for class ``racing car''. 
    The image on the left is a hard example for the class.     
    We show channel 909 for that image, at several resolutions and with GeM parameters $p^*=1$ and $p^*=3$.
   	In the low resolution version, the cars are too small to be visible individually on the activation map. 
	In the full resolution version, the location of the cars is more clear. 
	In addition, $p^*$\,$=$\,$3$ reduces the noisy detections relative to the true locations. 
}
\vspace{-7pt}
\end{figure}

\paragraph{Pooling exponent.} During the end-to-end training of our network, we consider two settings for the pooling exponent in the GeM layer of \cref{sec:p-pooling}: we set either $p=1$ or $p=3$. 
$p=1$ corresponds to average pooling, as used in classification architectures. 
The relevant literature~\cite{radenovic2018fine} and our preliminary experiments on off-the-shelf classification networks suggest that the value $p=3$ improves the retrieval performance on standard benchmarks. 
\Cref{fig:heatmap} illustrates this choice. 
By setting $p=3$, the car is detected with high confidence and without spurious detections. 
Boureau \etal~\cite{Boureau2010ATA} analyse average- and max-pooling of sparse features. 
They find that when the number of pooled features increases, it is beneficial to make them more sparse, which is consistent with the observation we make here.

\paragraph{Input size and cropping.} As described in~\cref{sec:input-size}, we train our network on crops of size $224\times 224$ pixels.
For testing, we experiment with computing MultiGrain embeddings at resolutions $s^*=224,500,800$. 
For resolution $s^*=224$, we follow the classical image classification protocol ``resolution 224'': the smallest side of an image is resized to $256$ and then a $224\times 224$ central crop is extracted.
For resolution $s^* > 224$, we instead follow the protocol common in image retrieval and resize the largest side of the image to the desired number of pixels and evaluate the network on the rectangular image, without cropping. 

\paragraph{Margin loss and batch sampling.} We use $m$\,$=$\,3 data-augmented repetitions per batch.
We use the default margin loss hyperparameters of~\cite{wu2017sampling} (details in supplementary~\ref{sec:training-hyperparam}). 
As in~\cite{wu2017sampling} the distance-weighted sampling is performed independently on each of the 4 GPUs used for training.

\paragraph{Datasets.} We train our networks on the ImageNet-2012 training set of 1.2 million images labelled into $1,\!000$ object categories~\cite{ILSVRC15}. 
Classification accuracies are reported on the $50,\!000$ validation images of this dataset.
For image retrieval, we report the mean average precision on the {\bf Holidays} dataset~\cite{Jegou2008HammingEA}, with images rotated manually when necessary, as in prior evaluations on this dataset~\cite{Gordo2016DeepIR}. 
We also report the accuracy on the {\bf UKB} object recognition benchmark~\cite{nister2006scalable}, which contains $2,\!550$ instances of objects under $4$ varying viewpoints each; each image is used as a query to find its 4 closest neighbors in embedding space; the number of correct neighbors is averaged across all images, yielding a maximum score of $4$. 
We also report the performance of our network in a copy detection setting, indicating the mean average precision on the ``strong'' subset of the INRIA Copydays dataset~\cite{Douze2009EvaluationOG}.
We add 10K distractor images randomly sampled from the YFCC100M large-scale collection of unlabelled images~\cite{Thomee2016YFCC100MTN}.
We call the combination {\bf C10k}.

The PCA whitening transformations are computed from the features of 20K images from YFCC100M, distinct from the C10k distractors.

\subsection{Expanding resolution with pooling exponent\label{sec:expand-pooling}}

\begin{figure}
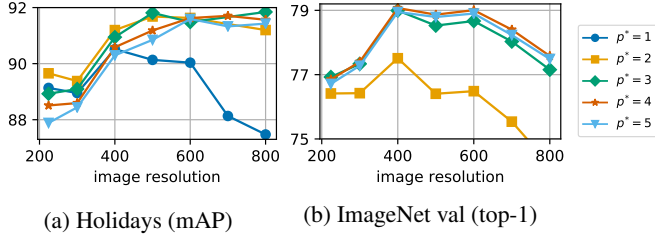

    \centering
    \begin{subfigure}[t]{0.42\columnwidth}
        \includegraphics[trim={0.47cm 0 0 0},clip,height=0.7\textwidth]{{{figs/p_plot/p_joint_3B_1.0-holidays}}}
        \vspace{-10pt}
        \caption{Holidays (mAP)}
        \label{fig:p_scale_holidays}
    \end{subfigure}\hfill%
    ~ %
    \begin{subfigure}[t]{0.42\columnwidth}
        \includegraphics[trim={0.47cm 0 0 0},clip,height=0.7\textwidth]{{{figs/p_plot/p_joint_3B_1.0-classif}}}
        \caption{ImageNet val (top-1)}
        \label{fig:p_scale_classif}
    \end{subfigure}%
    \begin{subfigure}[t]{0.13\columnwidth}
        ~\hspace{0.2cm}
        \raisebox{0.25in}[0pt][0pt]
        {\includegraphics[width=\textwidth]{{{figs/p_plot/legend}}}}
    \end{subfigure}%
\vspace{-3pt}
    \caption{\label{fig:p_pooling}
    	Retrieval and classification accuracies as a function of pooling exponent $p^*$ and the image resolution. 
    	At training time, the pooling was $p=3$. %
	Note the clear interaction between the resolution $s^*$ and the pooling exponent~$p^*$. 
    }
\vspace{-5pt}
\end{figure}

As our reference scheme, we train the network at resolution 224x224 with RA sampling and pooling exponent $p=3$. When testing on images with the same 224x224 resolution, this gives a $76.9\%$ top-1 validation accuracy on Imagenet, $0.7$\% points above the non-RA baseline, see \cref{tab:classifres}.

We now feed larger images at test time, \ie, we consider resolutions $s^*$\,$>$\,$224$ and vary the exponents $p^*$\,$\ne$\,$p$\,$=$\,$3$ at test time. 
\Cref{fig:p_scale_classif,fig:p_scale_holidays} show the classification accuracy on ImageNet validation and the retrieval accuracy on Holidays at different resolutions, for different values of the test pooling exponent $p^*$. 
As expected, at $s^*$\,=\,$224$, the pooling exponent yielding best accuracy in classification is the exponent with which the network has been trained, $p^*$\,=\,$3$. 
Observe that testing at larger scale requires an exponent $p^*$\,$>$\,$p$, both for classification and for retrieval. 

In the following, we adopt the values obtained by our cross-validation on \inaug, see \cref{sec:expanding-resolution}. 

\subsection{Analysis of the tradeoff parameter \label{sec:tradeoff-parameter}}

\begin{figure}[t]
\centering
\begin{minipage}{0.5\linewidth}
\includegraphics[width=\linewidth]{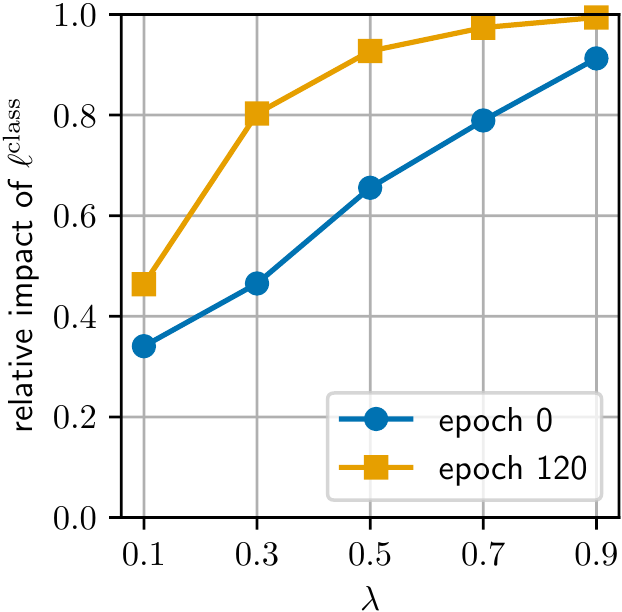}
\end{minipage}
\hfill
\begin{minipage}{0.45\linewidth}
\caption{\label{fig:gradients}
    Fraction of the classification and retrieval terms, measured as $\|g^\mathrm{class}\| / (\|g^\mathrm{class}\| + \|g^\mathrm{retr}\|)$, where the $g^\mathrm{class}$ vector is the gradient from the $\lambda \lclass$ component.
    Note, how the retrieval loss' influence is decreasing over epochs. 
    }
\end{minipage}
\vspace{-5pt}
\end{figure}

We now analyze the impact of the tradeoff parameter $\lambda$. 
Note, this parameter does not directly reflect the relative importance of the two loss terms during training, since these are not homogeneous: $\lambda$\,=\,0.5 does not mean that they have equal importance. \Cref{fig:gradients} analyzes the actual relative importance of the classification and margin loss terms, by measuring the average norm of the gradient back-propagated through the network at epochs 0 and 120. One can see that $\lambda$\,=\,$0.5$ means that the classification has slightly more weight at the beginning of the training. The classification term becomes dominant at the end of the training, meaning that the network has already learned to cancel data augmentation. 

In terms of performance, $\lambda$\,=$\,0.1$ leads to a poor classification accuracy. 
Interestingly, the classification performance is higher for the intermediate $\lambda$\,=\,$0.5$ ($77.4\%$ at $s^*$\,=\,$224$) than for $\lambda$\,=\,$1$, see \Cref{tab:classifres}. 
Thus, the margin loss leads to a performance gain for the classification task. 

We set $\lambda$\,$=$\,$0.5$ in our following experiments, as it gives the best classification accuracy at the practical resolutions $s^*$\,$=$\,$224$ and $500$ pixels. 
As a reference, we also report a few results with $\lambda$\,$=$\,$1$. 

\subsection{Classification results \label{sec:classif-results}}

From now on, our MultiGrain nets are trained at resolution $s$\,=\,$224$ with exponent $p$\,=\,$1$ (standard average pooling) or $p$\,=\,$3$ in the GeM pooling. 
For each evaluation resolutions $s^*$\,=\,$224, 500, 800$, the same exponent $p^*$ is selected according to \cref{sec:expanding-resolution}, yielding a single embedding for classification and for the retrieval. 
\Cref{tab:classifres} presents the classification results. 
There is a large improvement in classification performance from 
our baseline Resnet-50 with $p$\,=\,$1$, $s$\,=\,$224$, ``full'' data augmentation (76.2\% top-1 accuracy), 
to a MultiGrain model at $p$\,=\,$3$, $\lambda$\,=\,$0.5$, $s$\,=\,$500$ (78.6\% top-1).
We identify four sources for this improvement:
\begin{enumerate}
\item Repeated augmentations: adding RA batch sampling (\cref{sec:data-augmented-batches}) yields an improvement of $+0.6 \%$ $(p$\,$=$\,$1)$.
\item Margin loss: the retrieval loss helps the generalizing effect of data augmentation: $+0.2 \%$ $(p$\,$=$\,$1)$.
\item $p$\,$=$\,$3$ pooling: GeM at training (\cref{sec:p-pooling}) allows the margin loss to have a much stronger effect thanks to increased localization of the features: $+0.4\%$.
\item Expanding resolution: evaluating at resolution $500$ adds $+1.2\%$ to the $p$\,$=$\,$3$ MultiGrain network, reaching the $78.6$ top-1 accuracy. 
This is made possible by the $p$\,$=$\,$3$ training -- which yields sparser features, more generalizable over different resolutions, and by the $p^*$ pooling adaptation -- without it the performance at this resolution is only $78.0\%$.
\end{enumerate}

The $p^*$ selection for evaluation at higher resolutions has its limits: at $800$ pixels, due to the large discrepancy between the training and testing scale for the feature extractor, the accuracy drops to $77.2\%$ ($76.2\%$ without the $p^*$ adaptation).

\begin{table}[t]
\centering
\caption{\label{tab:classifres}
	ImageNet 2012 validation performance at top-1~/~top-5 accuracies ($\%$). 
    Resnet-50 is a %
    classification baseline trained with cross-entropy with our training schedule, data augmentation, and uniform batch sampling. 
    MultiGrain uses the same Resnet-50 trunk. 
At resolutions $s^*$\,$>$\,224 we evaluate with exponent $p^*$ as described in \cref{sec:expanding-resolution}. %
}
\vspace{-2pt}
{\small
\begin{tabular*}{\columnwidth}{@{\extracolsep{\fill}}lcccccc}
\toprule
Architecture & $\lambda$ & data & resol. & \multicolumn{2}{c}{train-time pooling}\\ 
& & aug. & $s^*$ & $p = 1$ & $p = 3$ \\
\midrule

ResNet-50  & & full & $224$ & $76.2$ / $92.9$ & $76.2$ / $93.1$\\
MultiGrain & $1$ & full & $224$ & $76.8$ / $93.2$ & $76.9$ / $93.5$ \\
MultiGrain & $0.5$ & full & $224$ & $77.0$ / $93.6$ & $\bm{77.4}$ / $\bm{93.6}$ \\
MultiGrain & $0.5$ & AA & $224$ & $77.4$ / $93.6$ & $\bm{78.2}$ / $\bm{93.9}$ \\
MultiGrain & $0.5$ & full & $500$ & $76.5$ / $93.5$ & $\bm{78.6}$ / $\bm{94.4}$ \\
MultiGrain & $0.5$ & AA & $500$ & $77.7$ / $94.0$ & $\bm{79.4}$ / $\bm{94.8}$ \\
MultiGrain & $0.5$ & full & $800$ & $73.5$ / $93.5$ & $77.2$ / $93.5$ \\
MultiGrain & $0.5$ & AA & $800$ & $74.1$ / $91.8$ & $77.8$ / $93.9$ \\
\midrule
\multicolumn{3}{l}{PyTorch model zoo} & $224$ & $76.1$ / $92.9$ \\
\multicolumn{3}{l}{mixup~\cite{zhang2018mixup}} & $224$ & $76.7$ / $94.4$ \\
\multicolumn{3}{l}{BA ($|\mathcal{B}| = 1024$)~\cite{2019arXiv190109335H}} & $224$ & $76.9$ / \makebox[\widthof{$94.5$}][c]{--} \\
\multicolumn{3}{l}{AutoAugment~\cite{cubuk2018autoaugment}} & $224$ & $77.6$ / $93.8$ \\
\bottomrule
\end{tabular*}

}
\vspace{-2pt}
\end{table}

\paragraph{AutoAugment}~\cite{cubuk2018autoaugment} (AA) is a method to learn data-augmentation  
using reinforcement learning techniques to improve the accuracy of classification networks on ImageNet. 
We %
directly integrate the data-augmentations found by the algorithm 
\cite{cubuk2018autoaugment} trained on their Resnet-50 model using a long schedule of 270 passes over the dataset, with batch size $4096$. 
We have observed that this longer training gives more impact to the AA-generated augmentations. 
We therefore use a longer schedule of 7508 iterations per epoch, keeping the batch size to $|\mathcal{B}|=512$. 

Our method benefits from this data-augmentation: 
MultiGrain reaches $78.2\%$ top-1 accuracy at resolution $224$ with $p$\,$=$\,3, $\lambda$\,$=$\,0.5.
To the best of our knowledge, this is the best top-1 accuracy reported for Resnet-50 when training and evaluating at this resolution, significantly higher than the $77.6\%$ reported with AutoAugment alone~\cite{cubuk2018autoaugment} or $76.7\%$ for mixup~\cite{zhang2018mixup}.
Using a higher resolution at test time improves the accuracy further: we obtain $79.4\%$ top-1 accuracy at resolution $500$. 
Our strategy of adapting the pooling exponent to a larger resolution is still effective, and significantly outperforms the state of the art performance for a ResNet-50 learned on ImageNet at training resolution $224$. %

\begin{table}
\centering
\caption{\label{tab:instanceres}
	Instance search results and baselines, on Holidays ($\%$~mAP) and UKB ($/4$).
	We set $p=3$ pooling at training time for our MultiGrain models, and $p^*$ set as given in \cref{sec:expanding-resolution}. 
	$\dagger${\small \em GeM is fine-tuned at resolution 362x362 on additional images tailored to the retrieval task. Their best result is obtained with multi-scale input and implies additional processing.
}}
\vspace{-2pt}

\def \mysp {\hspace{3pt}}
{\small
\begin{tabular}{@{\hspace{5pt}}l@{\hspace{5pt}}r ccc@{\hspace{5pt}}}
\toprule
Method & resol. $s^*$ & Holidays & UKB & CD10k \\ 
\midrule
MultiGrain $\lambda=1$                     & 500  & {\bf 91.8} & 3.89       & {\bf 81.1} \\
MultiGrain $\lambda=1$                     & 800  & 91.6       & {\bf 3.91} & {\bf 82.5} \\
MultiGrain $\lambda=0.5$                   & 500  & 91.5       & {\bf 3.90} & 80.7 \\
MultiGrain $\lambda=0.5$                   & 800  & {\bf 92.5} & {\bf 3.91} & 78.6 \\
\hline 
Fisher vectors~\cite{jegou2012aggregating} & 800  & 63.4       & 3.35       & 42.7 \\
Neural codes~\cite{babenko2014neural}      & 224  & 79.3       & 3.56 \\
ResNet-50 RMAC~\cite{Gordo2016DeepIR}      & 724  & 90.9  \\
ResNet-50 RMAC~\cite{Gordo2016DeepIR}      & 1024 & 93.3 \\
ResNet-101 RMAC~\cite{Gordo2017EndtoEndLO} & 800  & 91.4 & 3.89\\
GeM$^\dagger$~\cite{radenovic2018fine}     & {\bf 1024} &  {\bf 93.9} \\
\bottomrule
\end{tabular}
}
\vspace{-5pt}
\end{table}

\subsection{Retrieval results \label{sec:retrieval-results}}

We present our retrieval results in \cref{tab:instanceres}, with an ablation study and copy-detection results in the supplemental material (\ref{sec:ablationretrieval}). 
Our MultiGrain nets improve accuracies on all datasets with respect to the Resnet-50 baseline for comparable resolutions. 
Repeated augmentations (RA) is again a key ingredient in this context.

We compare with baselines where no annotated retrieval dataset is used. 
\cite{Gordo2016DeepIR,Gordo2017EndtoEndLO} give off-the-shelf network accuracies with R-MAC pooling. 
MultiGrain compares favorably with their results at a comparable resolution ($s^*$\,$=$800). 
They reach accuracies above $93\%$ mAP on Holidays but this requires a resolution $s$\,$\ge$\,1000 pixels. %

It is also worth noting that we reach reasonable retrieval performance at resolution $s^*$\,$=$\,500, which is a interesting operating point with respect to the traditional inference resolutions $s$\,$=$\,$800$--$1000$ for retrieval. 
Indeed, a forward pass of Resnet-50 on 16 processor cores takes $3.80$s at resolution $500$, against $18.9$s at resolution $1024$ ($5\times$ slower).
Because of this quadratic increase in timing, and the single embedding computed by MultiGrain, our solution is particularly adapted to large-scale or low-resource vision applications.

For comparison, we also report some older related results on the UKB and C10k datasets, that are not competitive with MultiGrain. 
Neural codes~\cite{babenko2014neural} is one of the first works on retrieval with deep features. 
The Fisher vector~\cite{jegou2012aggregating} is a pooling method that uses local SIFT descriptors. 

At resolutions $500$ we see that the results with the margin loss ($\lambda$\,$=$\,0.5) are slightly lower than without ($\lambda$\,$=$1). 
This is partly due to the limited transfer from the \inaug task to the variations observed in retrieval datasets.

\section{Conclusion}\label{sec:conc}
In this work we have introduced 
MultiGrain, a unified embedding for image classification and instance retrieval.
MultiGrain relies on a classical convolutional neural network trunk, with a GeM layer topped with two heads at training time.
We have discovered that by adjusting this pooling layer we are able to increase the resolution of images used a inference time, while maintaining a small resolution at training time. 
We have shown that MultiGrain embeddings can perform well on classification and retrieval. 
Interestingly, MultiGrain also sets a new state of the art on pure classification compared to all results obtained with the same convolutional trunk.
Overall, our results show that retrieval and classification tasks can benefit from each other. 

An implementation of our method is open-sourced at \url{https://github.com/facebookresearch/multigrain}.

\ificcvfinal
\paragraph*{Acknowledgments.} 
We thank Kaiming He for useful feedback and references. 
Maxim Berman is supported by Research Foundation - Flanders (FWO) through project number G0A2716N. PSI--ESAT acknowledges a GPU server donation from FAIR Partnership Program.

\fi

{\small\bibliographystyle{ieee}\bibliography{biblio}}

\clearpage
\counterwithin{figure}{section}
\counterwithin{table}{section}
\counterwithin{equation}{section}
\appendix

\pagenumbering{Roman}  %
\pagestyle{plain}

\twocolumn[
  \begin{@twocolumnfalse}
\newpage
\null
\vskip .375in
\begin{center}
  {\Large \bf \inserttitle \\ \vspace{0.5cm} \large Supplementary Material \par}
  \vspace*{24pt}
  {
  \par
  }
\end{center}
\end{@twocolumnfalse}
]

We report a few additional experiments and results that did not fit in the main paper. 
Section~\ref{sec:suppl-data-augmented-toy} shows the effect of data-augmented batches when training a simple toy model.
Sections~\ref{sec:training-hyperparam} and \ref{sec:full-data-augment} list the values of a few hyper-parameters used in our method. 
Section~\ref{sec:ablationretrieval} gives a some more ablation results in the retrieval setting.
Finally, Section~\ref{sec:extra-classif} shows how to use the ingredients of MultiGrain to improve the accuracy of an off-the-shelf pre-trained ConvNet at almost no additional training cost. 
It obtains what appear to be \textbf{the best reported classification results on imagenet-2012} for a convnet with publicly available weights.

\section{Data-augmented batches: toy model \label{sec:suppl-data-augmented-toy}}
We have observed in \Cref{sec:data-augmented-batches,sec:tradeoff-parameter} that training our  architecture (ResNet-50 trunk) with data-augmented batches yields improvements with respect to the vanilla uniform sampling scheme, despite the decrease in  image diversity. 

\begin{figure}
\includegraphics[width=\linewidth]{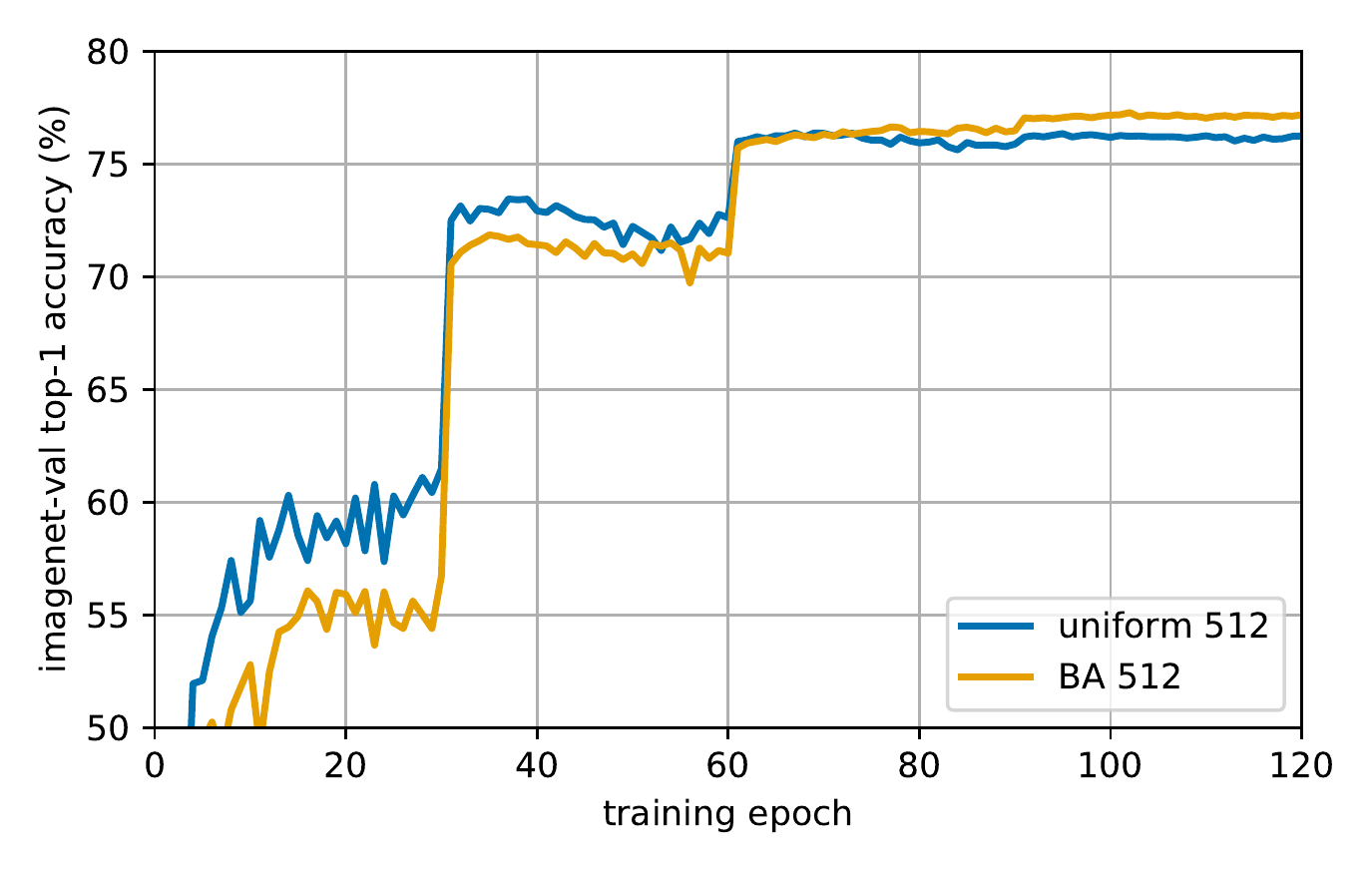}
\caption{\label{fig:suppl-compare-sampling}
	Evolution of the validation accuracy on ImageNet-val with and without data-augmented batches. 
}
\end{figure}

This observation holds even in the absence of ranking triplet loss, all things being equal otherwise: same number of iterations per epoch, number of epochs, learning rate schedule, and batch size. 
As an example, \cref{fig:suppl-compare-sampling} shows the evolution of the validation accuracy of our network trained under cross-entropy with our training schedule and a $p = 1$ pooling, batches of size 512, with the data augmentation introduced in \cref{sec:experimental-settings}, with uniform batches vs. with batch sampling. 
While initial epochs suffer from the reduced diversity of the batches compared to the uniformly-sampled variant, the reinforced effect on data augmentation compensates for this in the long run, and makes the batch-augmented variant reach a higher final accuracy. 

Since we observe this better performance even for a pure image classification task, an interesting question is whether this benefit is specific to our architecture and training method (batch-norm, etc), or if it is more generally applicable? 
Hereafter we analyse a linear model and synthetic classification task that seems to align with the second hypothesis.

\begin{figure}
\centering
\includegraphics[height=1.5in]{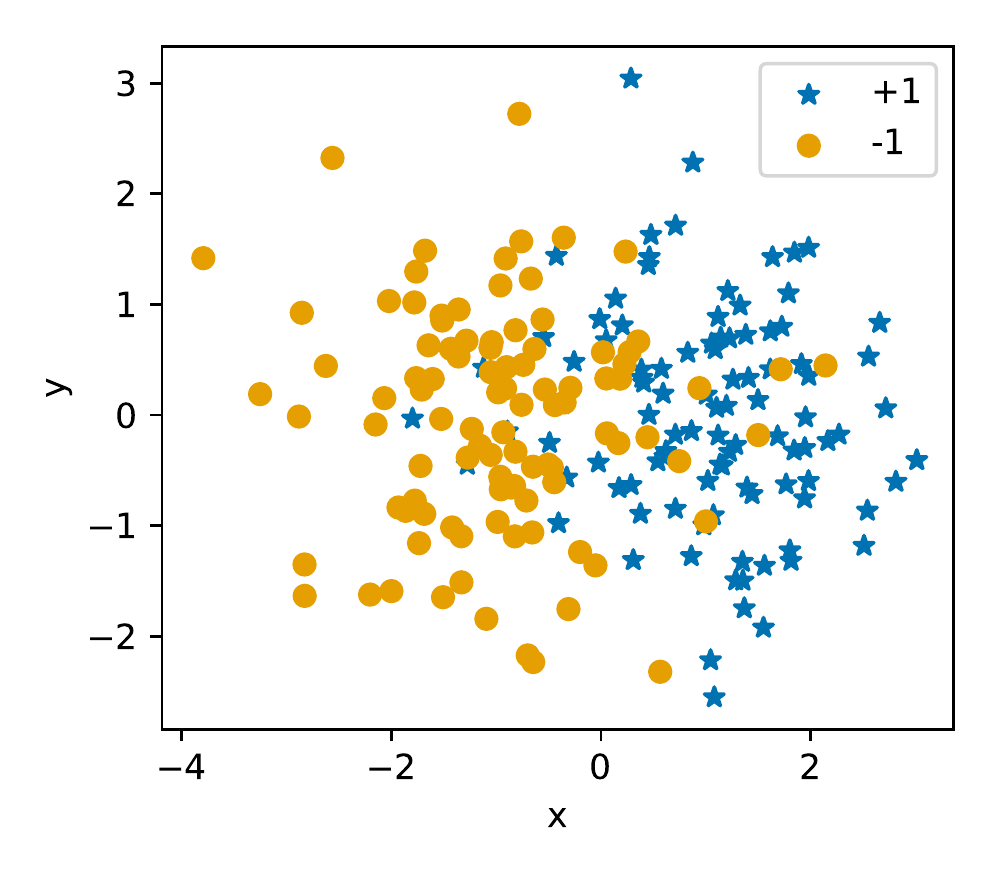}
\caption{\label{fig:toy-train}
	Training set for the toy model in~\cref{sec:suppl-data-augmented-toy}.
}
\end{figure}

We consider an idealized model of the effect of including different data-augmented instances of the same image in one batch using standard stochastic gradient descent. 
We create a synthetic training set $\mathcal{D}$ of points pictured in \cref{fig:toy-train} of $N$\,=\,$100$ positive and $N$\,=\,$100$ negative training points $\vec{p^i} = (p^i_x, p^i_y)$ by sampling from two 2D Gaussian distributions: 
\begin{equation}
\begin{aligned}
    p_x^i \sim \mathcal{N}(\mu=0, \sigma=1) \\
    p_y^i \sim \mathcal{N}(\mu=y_i^*, \sigma=1)
\end{aligned}
\end{equation}
with $y_i^* = \pm 1$ being the ground truth label. 
We sample a test dataset in the same manner. 

We consider the SGD training of an SVM
\begin{equation}\label{eq:svm-model}
    f_{\vec{w}}(\vec{p_i}) = \vec{w}^{\top} \vec{p_i}
\end{equation}
using the Hinge loss
\begin{equation}
    \ell^{\text{hinge}} = \max{(1 - y_i^* f_{\vec{w}}(\vec{p_i}), 0)}.
\end{equation}

We consider the symmetry across the x-axis
\begin{equation}
    \phi((p^i_x, p^i_y)) = \phi((p^i_x, -p^i_y))
\end{equation}
as a label-preserving data-augmentation suited to our synthetic dataset. 
We train the SVM~\eqref{eq:svm-model} using one pass through the data-augmented dataset $\bar{\mathcal{D}}$ of size $4N$, using batches of size $2$.

\begin{figure}
\includegraphics[width=\linewidth]{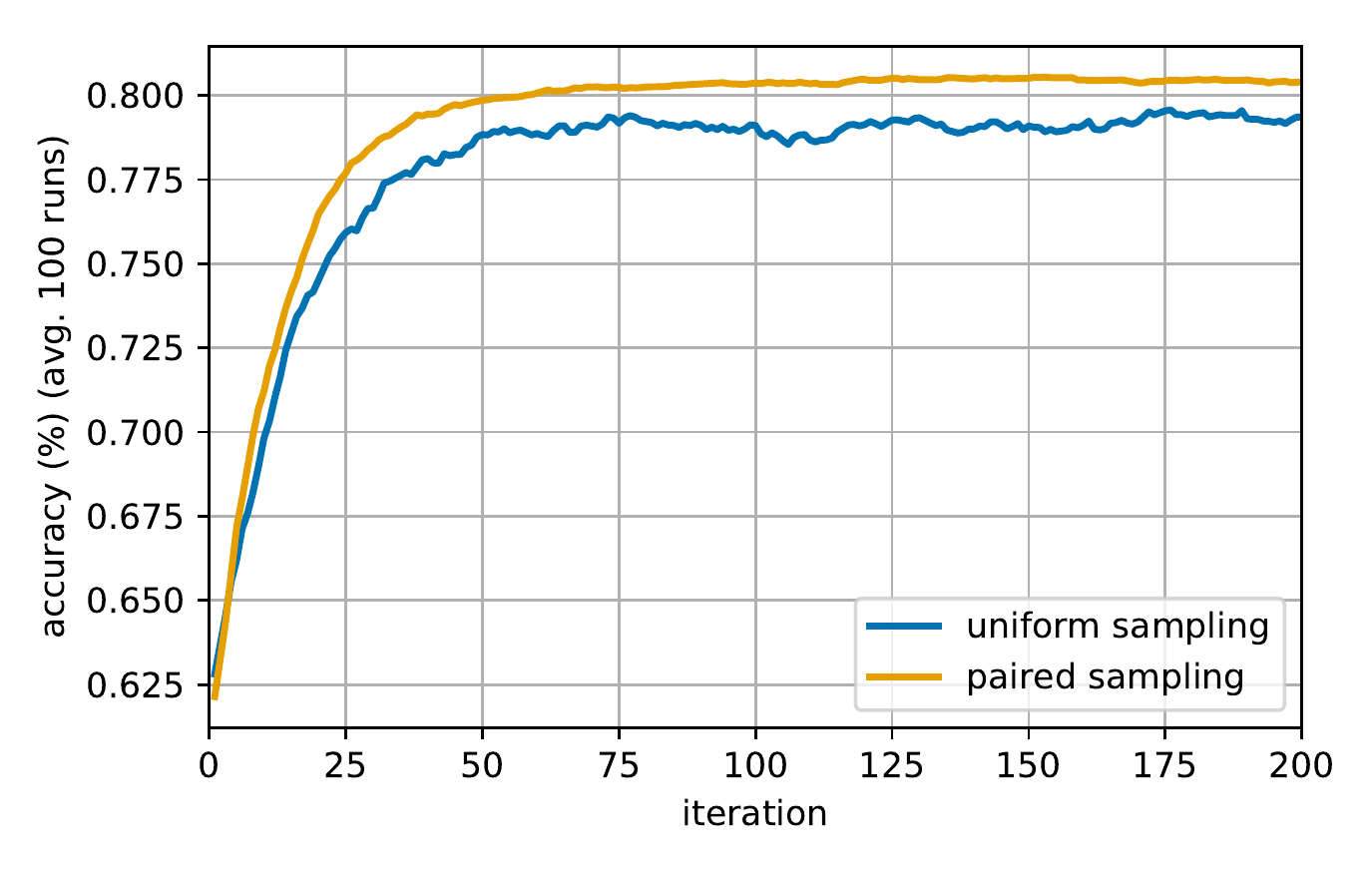}
\caption{\label{fig:evo-toy}
	Evolution of the test accuracy of the SVM trained on the synthetic data, averaged accross 100 runs.
}
\end{figure}

The only difference between the two optimization schedules is the order in which the samples are batched and presented to the optimizer. 
We consider two batch sampling strategies:
\begin{itemize}
    \item Uniform sampling: we sample the elements of the batch randomly from  $\bar{\mathcal{D}}$, without replacement;
    \item Paired sampling: we generate a batch by pairing a random element from $\bar{\mathcal{D}}$ and its data-augmentation, removing these two elements from $\bar{\mathcal{D}}$.
\end{itemize}
\Cref{fig:evo-toy} shows the evaluation of the accuracy with the iterations in both of these cases, averaged across 100 runs. 
It is clear that pairing the data-augmented pairs in one batch accelerates the convergence of this model. 

This idealized experiment demonstrates that there are cases in which the repeated augmentation scheme provides an optimization and generalization boost, and reinforces the effect of data augmentation.

\section{Margin loss hyper-parameters}

\label{sec:training-hyperparam}

\Cref{tab:training-hyperparam} 
gives the value of the hyper-parameters for the margin loss used during the training of our models.

\begin{table}
\centering
\caption{\label{tab:training-hyperparam}
	Margin loss hyper-parameters
}
\begin{tabular}{@{}lc@{}}
\toprule
parameter             & value \\ \midrule
margin $\alpha$       & $0.2$ \\
initial $\beta_0$     & $1.2$ \\
$\beta$ learning rate & $0.1$ \\
\bottomrule
\end{tabular}
\end{table}

\section{Data augmentation hyper-parameters}

\label{sec:full-data-augment}

\begin{table}
\centering
\caption{\label{tab:full-data-augment}
	\emph{full} data-augmentation transforms and parameters
}
{\small
\begin{tabular}{@{}lc@{}}
\toprule
\textbf{transformation} & \textbf{parameter range} \\ \midrule
horizontal flip           & \\
\midrule
random resized crop       & \makecell{$\text{scale}\in [0.08, 1.0]$ \\ $\text{ratio}\in [3/4, 4/3]$} \\
\midrule
color jitter & \makecell{brightness $0.3$\\contrast $0.3$\\
saturation $0.3$}\\
\midrule
lighting transform & intensity $0.1$ \\
\bottomrule
\end{tabular}
}
\end{table}

\Cref{tab:full-data-augment} gives the transformations in the \emph{full} data augmentation used in our experiments (\cref{sec:experimental-settings}), along with their parameters. 

\begin{table*}[t]
\renewcommand\thetable{D.1}
\centering
\caption{\label{tab:instanceres-abl}
	Full results including Copydays + 10k distractors (CD10k, $\%$~mAP), and ablation study for the MultiGrain models. 
	The Pytorch model simply extract the last activation layer as a descriptor~\cite{babenko2014neural}. 
	Resnet-50 corresponds to features extracted from a 
    classification baseline with $p = 1$ or $p = 3$ GeM pooling, trained with cross-entropy with our training schedule, data augmentation, and uniform batch sampling.
}
\def \mysp {\hspace{5pt}}
{\small
\begin{tabular}{lc c c@{\mysp}c@{\mysp}c c c@{\mysp}c@{\mysp}c c c@{\mysp}c@{\mysp}c}
\toprule
& & & \multicolumn{3}{c}{Holidays} & &  \multicolumn{3}{c}{UKB} & & \multicolumn{3}{c}{CD10k} \\
\cmidrule(rl){4-6} \cmidrule(rl){8-10} \cmidrule(rl){12-14}
Method & $\lambda$ & $s^*=$ & 224 & 500 & 800 &  & 224 & 500 & 800 &  & 224 & 500 & 800 \\
\midrule
\multicolumn{2}{l}{PyTorch model zoo} & & $85.5$ & $86.6$ & $82.8$ &      & $3.71$ & $3.85$ & $3.80$ &    & $61.5$ & $61.1$ & $43.0$\\
\multicolumn{2}{l}{Resnet-50 trained with $p = 1$ pooling} & & $83.5$ & $88.8$ & $87.1$ &    & $3.60$ & $3.79$ & $3.82$ &    & $59.2$ & $69.9$ & $66.2$\\
\multicolumn{2}{l}{Resnet-50 trained with $p = 3$ pooling} & & $86.8$ & $90.0$ & $90.4$ &    & $3.73$ & $3.87$ & $3.89$ &    & $70.6$ & $78.9$ & $75.7$ \\
MultiGrain & $1$ & & $\bm{88.9}$ & $\bm{91.8}$ & $91.6$ &  & $\bm{3.78}$ & $3.89$ & $\bm{3.91}$ & & $\bm{75.1}$ & $\bm{81.2}$ & $\bm{82.5}$ \\
MultiGrain & $0.5$ & & $88.3$ & $91.5$ & $\bm{92.5}$ & & $\bm{3.78}$ & $\bm{3.90}$ & $\bm{3.91}$ & & $74.1$ & $80.7$ & $78.6$  \\
MultiGrain + AA & $0.5$ & & $86.5$ & $90.3$ & $89.4$ & & $3.75$ & $3.89$ & $3.90$ & & $69.7$ & $77.8$ & $76.1$  \\
\bottomrule
\end{tabular}

}
\end{table*}

\section{Additional results and ablation study for Multigrain in retrieval}
\label{sec:ablationretrieval}

\begin{table*}[tb]
\renewcommand\thetable{E.1}
\centering
\caption{\label{tab:extra-classif}
    Additional top-1/top-5 validation classification accuracies obtained by finetuning $p^*$ for higher evaluation scales on off-the-shelf networks.
    The first column indicates the training resolution $s$ and the accuracy we measured at this resolution, with standard evaluation (resize of the largest scale to $s \cdot 256/224$ + center crop). 
    The subsequent columns show the accuracy measured at higher resolutions $s^*=350,400,450,500$ without cropping, together with the $p^*$ found by finetuning for these resolutions ($\cref{sec:extra-classif}$).
}
\begin{tabular}{@{}lcccccccccc@{}}
\toprule
 & \multicolumn{2}{c}{original evaluation} &
 \multicolumn{2}{c}{$s^* = 350$} & \multicolumn{2}{c}{$s^* = 400$} &
 \multicolumn{2}{c}{$s^* = 450$} &
 \multicolumn{2}{c}{$s^* = 500$} \\ 
 \cmidrule(rl){2-3}\cmidrule(rl){4-5} \cmidrule(rl){6-7}\cmidrule(rl){8-9}
 \cmidrule(rl){10-11}
 Architecture & \multicolumn{1}{c}{$s$} & acc. (\%) & 
 $p^*$ &
 acc. (\%) & 
 $p^*$ &
 acc. (\%) & $p^*$ & acc. (\%) & $p^*$ & acc. (\%) \\ \midrule
NASNet-A-Mobile~\citesuppl{zoph2018learning}%
& 224 & $74.1$/$91.7$ & $1.7$ & $\bm{75.1}/\bm{92.5}$ & $2.1$ & $74.2$/$92.1$ & $2.4$ & $71.8$/$90.9$ & $2.6$ & $68.4/89.0$ \\
SENet154~\cite{hu2018squeeze} & $224$ &$81.3$/$95.5$ & $1.6$ & $82.6/96.2$ & $1.6$ & $83.0$/$96.5$ & $1.6$ & $\bm{83.1}/\bm{96.5}$ & $1.7$ & $82.7$/$96.3$ \\
PNASNet-5-Large~\citesuppl{liu2018progressive}%
& $331$ & $82.7$/$96.0$ & $1.0$ & $81.3$/$85.4$ & $1.4$ & $82.6$/$96.1$ & $1.5$ & $83.2$/$96.4$ & $1.7$ & $\bm{83.6}/\bm{96.7}$ \\
\bottomrule
\end{tabular}
\end{table*}

\Cref{tab:instanceres-abl} reports additional results of the MultiGrain architecture, with an ablation study analyzing the effect of each component. 

As already reported in the main paper, for some datasets the choice of not using the triplet loss ($\lambda=1$) is as good or better than our generic choice ($\lambda=0.5$). 
Of course, then the embedding is not multi-purpose anymore. 
Overall, the different elements employed in our architecture (RA and the layers specific to Multigrain) still give a significant improvement over simply using the activations, and is competitive with the state of the art for the same resolution/complexity. 

Note, the AutoAugment data augmentation does not transfer well to the retrieval tasks. 
This can be explained by their specificity to Imagenet classification. 
This shows the limitation of a particular choice of data-augmentation  if a single embedding for classification and retrieval datasets is desired. 
Learning AutoAugment specifically for the retrieval task would certainly help, but would probably also result in less general embeddings. Hence, data-augmentation is a limiting factor for multi-purpose embeddings: improving for one task like classification hurts the performance for other  tasks.

\section{Evaluation of off-the-shelf classifiers at higher resolutions\label{sec:extra-classif}}

In this section, we present some additional classification results using off-the-shelf pretrained classification networks trained with standard average pooling ($p=1$).

As outlined in~\cref{sec:expand-pooling,sec:expanding-resolution}, one of our contributions is a strategy for evaluating classifier networks trained with GeM pooling at scale $s$ and exponent $p$ at a higher resolution $s^*$ and adapted exponent $p^*$. 
It can be used on pretrained networks as well.

For an evaluation scale $s^*$, we use the alternative strategy described in \cref{sec:expanding-resolution} to choose $p^*$: 
we finetune the parameter $p^*$ by stochastic gradient descent, backpropagating the cross-entropy loss on training images from imagenet, rescaled to the desired input resolution. 
Compared to a full finetuning at this input resolution, this strategy has a limited memory footprint, given that the backpropagation only has to be done on the ultimate classification layer before reaching the pooling layer, allowing for an efficient computation of the gradient of $p^*$.
Experimentally we also found that this process converges on a few thousands of training samples, while a finetuning of the classification layer would  require several data-augmented epochs on the full training set.

The finetuning is done using SGD with batches of $|\mathcal{B}| = 4$ (non-cropped) images, with momentum $0.9$ and initial learning rate $\text{lr}^{(0)} = 0.005$, decayed under a polynomial learning rate decay
\begin{equation}
    \text{lr}^{(i)} = \text{lr}^{(0)} \left( 1 - \frac{i}{i_\text{max}} \right)^{0.9}
\end{equation}
with $i_\text{max}$ the total number of iterations.

We select $50,000$ images from the training set ($50$ per category) for the fine-tuning and do one pass on this reduced dataset. We use off-the-shelf pretrained convnets from the \emph{Cadene/pretrained-models.pytorch} GitHub repository\footnote{Url: \url{https://github.com/Cadene/pretrained-models.pytorch}}. 
Table~\ref{tab:extra-classif} outlines the resulting validation accuracies. 
We see that for each network there is a scale and choice of $p^*$ that performs better than the standard evaluation. 

These networks have not been trained using GeM pooling with $p>1$; as exhibited in our classification results (\cref{tab:classifres}) we found this to be another key ingredient in ensuring a higher scale insensitivity and better performance at larger resolution. 
As in our main experiments with the MultiGrain architecture with a ResNet-50 backbone,
it is likely that these networks would reach higher values when training from scratch with a $p>1$ pooling, and adding repeated augmentations and margin loss.
However, running training experiments on these large networks is significantly more expensive. 
Therefore, we leave this for future work.

\FloatBarrier
{\small\bibliographystylesuppl{ieee}\bibliographysuppl{biblio}}

\end{document}